\pdfoutput=1

\documentclass[11pt]{article}

\usepackage[]{acl}

\usepackage{times}
\usepackage{latexsym}

\usepackage{todonotes}

\usepackage[T1]{fontenc}

\usepackage[utf8]{inputenc}

\usepackage{microtype}

\usepackage[absolute,overlay]{textpos}

\usepackage{multirow}
\usepackage{arydshln}
\usepackage{graphicx}
\usepackage{booktabs}
\usepackage{amssymb}  
\usepackage{subcaption}
\usepackage{amsmath} 

\usepackage{diagbox}
\usepackage{subcaption}

\usepackage{color, colortbl}

\usepackage{pifont}

\usepackage{tabularx}
\usepackage{colortbl}
\usepackage{tikz}

\usepackage{xspace}
\usepackage{arydshln}

\title{Lifting the Curse of Multilinguality 
\\ by Pre-training Modular Transformers}

\author{Jonas Pfeiffer\thanks{\,\,\,\,Work done while interning at Meta AI.}$^{\,\,\,1,2,3}$, \quad Naman Goyal$^3$, \quad Xi Victoria Lin$^3$, \quad Xian Li$^3$,  \\ \textbf{ James Cross$^3$,  \quad Sebastian Riedel$^3$,  \quad Mikel Artetxe$^3$} \\
$^1$New York University, $^2$TU Darmstadt, \\
$^3$Meta AI
}

\begin{document}
\maketitle

\begin{abstract}

Multilingual pre-trained models are known to suffer from \textit{the curse of multilinguality}, which causes per-language performance to drop as they cover more languages. We address this issue by introducing language-specific modules, which allows us to grow the total capacity of the model, 
while keeping the total number of trainable parameters per language constant. In contrast with prior work that learns language-specific components post-hoc, we pre-train the modules of our \textbf{Cross}-lingual \textbf{Mod}ular (\textbf{\textsc{X-Mod}}) models  from the start. Our experiments on natural language inference, named entity recognition and question answering show that our approach not only mitigates the negative interference between languages, but also enables positive transfer, resulting in improved monolingual and cross-lingual performance. Furthermore, our approach enables adding languages post-hoc with no measurable drop in performance, no longer limiting the model usage to the set of pre-trained languages. 

\end{abstract}

\section{Introduction}
Recent work on multilingual NLP has focused on pre-training transformer-based models \cite{Vaswani2017transformer} on concatenated corpora of a large number of languages \cite{Devlin2019bert, Conneau2020xlm-r}. These multilingual models have been shown to work surprisingly well in cross-lingual settings, despite the fact that they do not rely on direct cross-lingual supervision  \cite[e.g., parallel  data or translation dictionaries;][]{Pires2019, wu-dredze-2019-beto,artetxe-etal-2020-cross,Hu2020xtreme, K2020Crosslingab, rust-etal-2021-good}.

\begin{figure}[!t]
    \centering
        \includegraphics[width=0.99\linewidth]{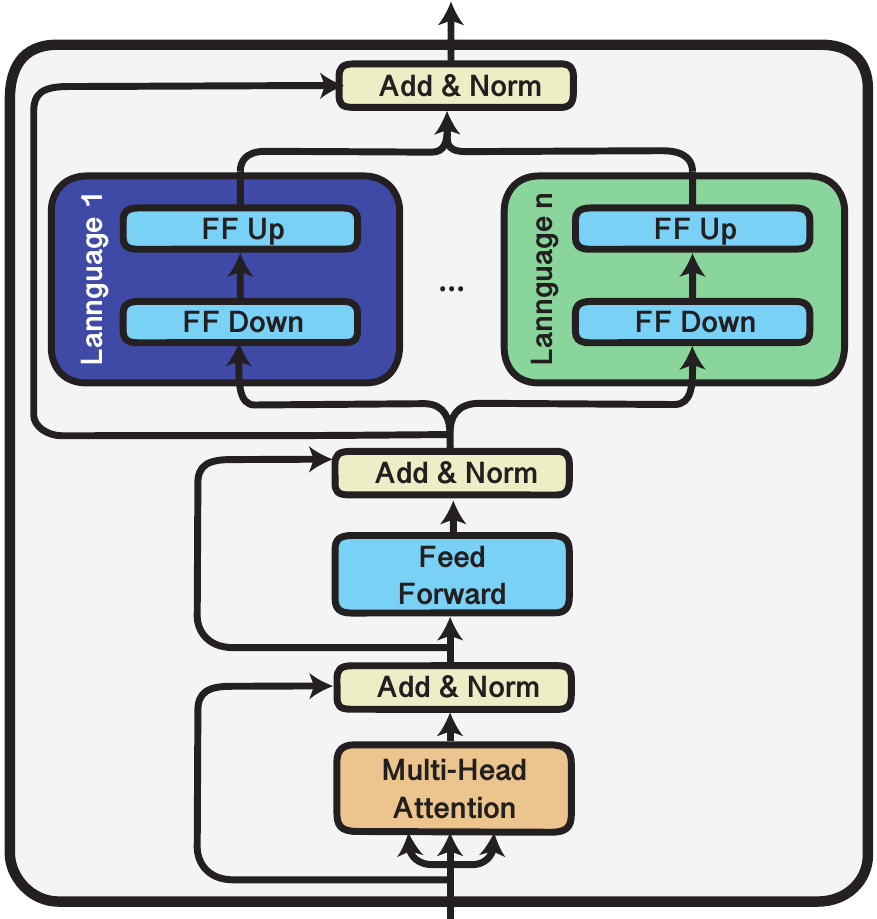}
    \caption{A transformer layer of our proposed modular architecture. The dark blue and green components illustrate the modular layers, which are language specific. The Multi-Head Attention and Feed-Forward components  are shared by all languages.
     }

\label{fig:modular_architecture}
\end{figure}

\begin{figure*}[!t]
    \centering
    
        \begin{subfigure}[]{0.45\textwidth}
        \includegraphics[width=0.99\textwidth]{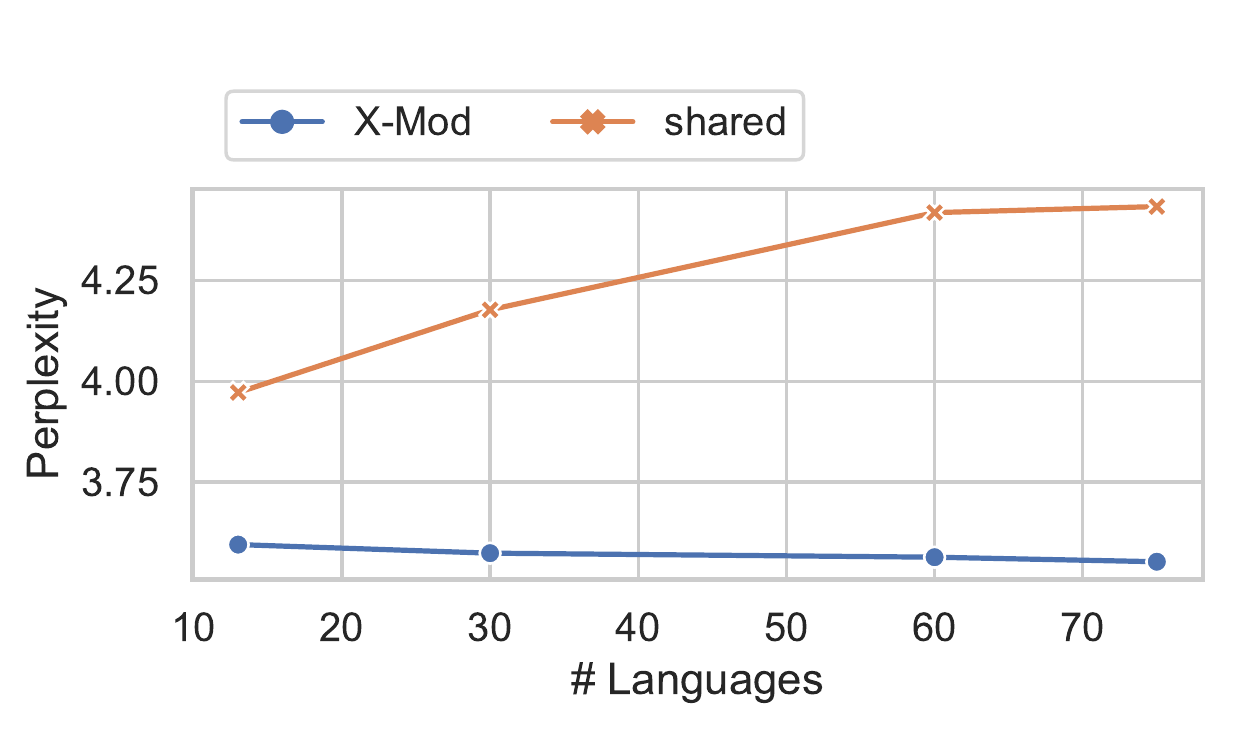}
        \caption{ Mean Perplexity. }
        \label{fig:Mean_Perplexity}
        
    \end{subfigure}
    \begin{subfigure}[]{0.45\textwidth}
        \includegraphics[width=0.99\textwidth]{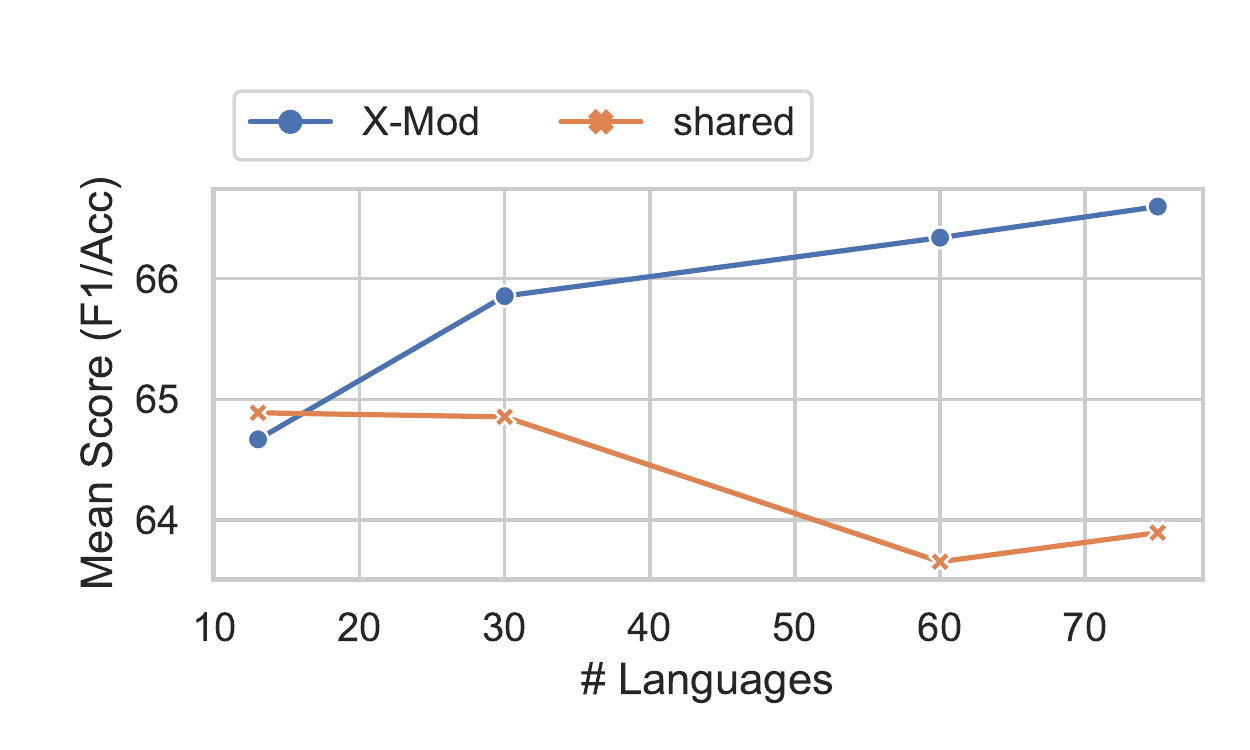}
        \caption{ Mean Performance on XNLI and NER. }
    \end{subfigure}
    
    \caption{ Average  (a) perplexity and (b) transfer performance on XNLI and NER  across pre-trained languages when training on an increasing number of  languages. Each model has seen the \textbf{same amount of examples} in each language. Lower perplexity and higher downstream score indicate better performance. Refer to Figure~\ref{fig:maintaining_update_steps} for per-task performance, and Appendix \ref{app:additional_results} for per-language performance. 
    }

\label{fig:perplexity_total_line}
\end{figure*}

However, recent work has uncovered fundamental limitations of  multilingual transformers. \citet{Conneau2020xlm-r} observe that pre-training a model with a fixed capacity on an increasing amount of languages only improves its cross-lingual performance up to a certain point, after which performance drops can be measured---a phenomenon known as \textit{the curse of multilinguality} (Figure~\ref{fig:perplexity_total_line}).
As such, prior work   had to find a trade-off between supporting more languages and obtaining better performance on a smaller set of languages.

In this work, we address this problem
by introducing language-specific, modular components during pre-training (Figure~\ref{fig:modular_architecture}). Our \textbf{Cross}-lingual, \textbf{Mod}ular (\textbf{\textsc{X-Mod}}) language model shares the majority of the transformer parameters between all pre-training languages, while providing each language with individual capacity to learn idiosyncratic information without increasing the total number of trainable parameters per language.
While previous adapter-based approaches (Figure~\ref{fig:mod_ada}) extend pre-trained multilingual language models (LMs) with modular components \textit{after} pre-training, we  add modular components  \textit{during} pre-training, thereby preparing the model to be extended to new languages post-hoc. 
Our experiments on natural language inference (NLI), named entity recognition (NER), and question answering (QA) demonstrate that  our modular architecture not only is effective at mitigating interference between languages, but also achieves positive transfer, resulting in improved monolingual and cross-lingual performance. In addition, we show that \textsc{X-Mod}  can be extended to unseen languages, with no measurable drop in performance, by learning its corresponding modules and leaving the shared parameters frozen.
All in all, we propose a multilingual architecture that can scale to a large number of languages without any loss in performance, and can be further extended to new languages after pre-training.\footnote{Code
and pre-trained models are available at: \href{https://github.com/pytorch/fairseq/tree/main/examples/xmod}{https://github.com/pytorch/fairseq/tree/main/examples/xmod}.}

\section{Background and related work}

We provide a background on multilingual and modular language modelling, as well as approaches that extend LMs to new languages.

\subsection{Multilingual transformers}

Recent LMs \cite{Devlin2019bert, Conneau2020xlm-r}, based on transformer architectures \cite{Vaswani2017transformer} and pre-trained on massive amounts of multilingual data, have surpassed (static) cross-lingual word embedding spaces \cite{Ruder:2019jair,Glavas:2019acl} 
for cross-lingual transfer in NLP \cite{Pires2019,wu-dredze-2019-beto,Wu2020emerging,Hu2020xtreme,K2020Crosslingab}. Transformer-based models are \textbf{1)}~pre-trained on textual corpora using Masked Language Modelling (MLM). 
They are then \textbf{2)}~fine-tuned  on labelled data of a downstream task in a \textit{source} language and  \textbf{3)}~directly applied to perform inference in a \textit{target} language \cite{Hu2020xtreme}.

\begin{figure}[t!]
    \centering
    \begin{subfigure}[!t]{0.45\columnwidth}
        \centering
        \includegraphics[width=0.99\linewidth]{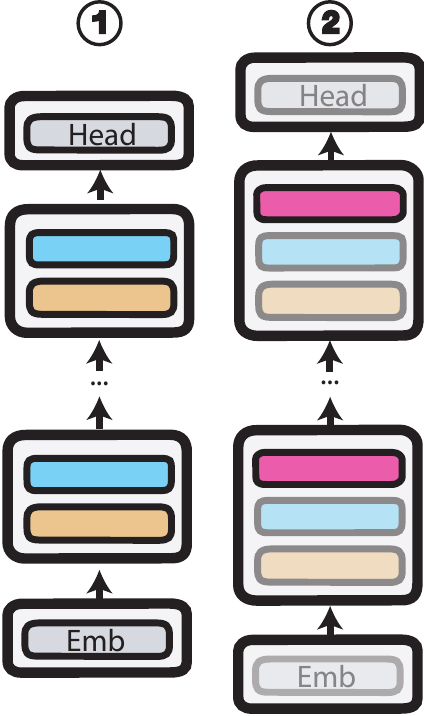}
        \caption{ Adapter-based}
        \label{fig:mod_ada}
    \end{subfigure}
        \hspace{5.0mm}
        \begin{subfigure}[!t]{0.45\columnwidth}
        \centering
        \includegraphics[width=0.99\linewidth]{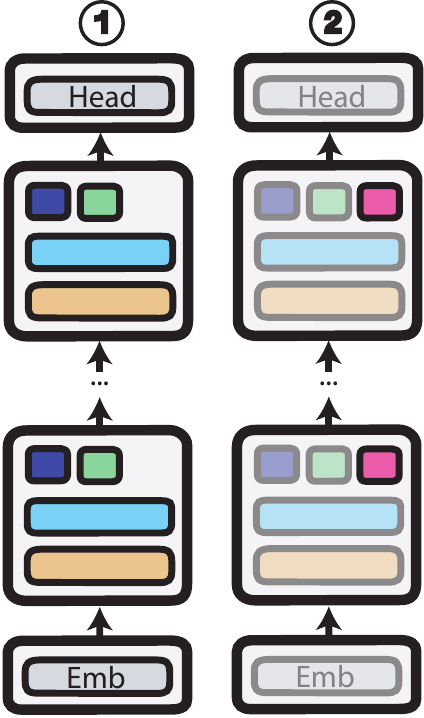}
        \caption{ \textsc{X-Mod}}
        \label{fig:mod_xmod}
    \end{subfigure}
    \caption{Our proposed architecture in comparison to adapter-based approaches.  (a)~Previous  approaches \ding{172}~utilize non-modular pre-trained transformer models  and \ding{173}~extend them with modular adapter components. (b)~We \ding{172}~pre-train the transformer with modular units from the get-go,  \textit{preparing} the model to be \ding{173}~extended with additional modular units later on. Yellow and light blue components indicate standard Multi-Head Attention and Feed-Forward layers. The remaining (non-gray) components are bottleneck (modular) units. Grayed-out components are frozen.}
    \label{fig:mod_ill}
\end{figure}

\subsection{Modular language models}

Modular approaches have a long standing history in NLP, preceding   pre-trained models \cite{Andreas2016}. They have recently re-gained interest for transformer-based models, where mixture of experts \cite[MoE;][]{Shazeer2017MoE} approaches have enabled training trillion parameters models in a distributed fashion \cite{Fedus2021}. More recently modular MoE approaches have been shown to improve domain-specific pre-training of LMs \cite{Gururangan2021Demix}. In a similar trend, `expert' modules have been added to (non-modular) pre-trained LMs post-hoc, predominantly referred to as adapters \cite{Rebuffi2017adapters, Rebuffi2018, Houlsby2019adapters}. Next to being extremely parameter  \cite{Houlsby2019adapters,  Mahabadi2021Compacter, He2021UnifiedAdapters} and training efficient \cite{pfeiffer2020AdapterHub, rueckle2020adapterdrop},  these modular approaches allow models to be extended to new data settings   \cite{chen2019slice,Ruckle2020}, where newly learned knowledge can be combined \cite{Stickland2019BertPals, Wang2021KAdapter,pfeiffer-etal-2021-adapterfusion, Lauscher2020comonsense, Mahabadi2020Hyper, Poth2021Pretrain}, or stacked for combinatory    cross-lingual \cite{pfeiffer-etal-2020-mad, pfeiffer2020unks,ustun-etal-2020-udapter, Vidoni2020OrthogonalLA, Ansell2021MADG, Ansell2021Composable, Wang2021Efficient} as well as NMT scenarios \cite{Bapna:2019emnlp, philip-etal-2020-monolingual, chronopoulou-etal-2020-reusing, Le:2021acl, Ustun2021DenoisingAda, Stickland2021DomainAdaNMT, garcia-etal-2021-towards}.

\subsection{Weaknesses, improvements, and extensions of language models}

Next to the \textit{curse of multilinguality}, recent works have shown substantially reduced cross-lingual and monolingual abilities of  models for low-resource languages with smaller pre-training data \cite{wu-dredze-2020-languages,Hu2020xtreme,Lauscher:2020zerohero,artetxe-etal-2020-cross,pfeiffer-etal-2020-mad,pfeiffer2020unks,ChauLS20Parsing,ponti-etal-2020-xcopa}.

\citet{K2020Crosslingab,artetxe-etal-2020-cross} show that a shared vocabulary is not necessary for cross-lingual transfer. 
\citet{Chung2021RethinkEmbCoup} demonstrate that decoupling the input embeddings from the prediction head improves the performance on a number of downstream tasks. \citet{dufter-schutze-2020-identifying} show that the number
of parameters and training duration is interlinked with the model's multilingual capability.  \citet{Chung2020Improving, rust-etal-2021-good} show that the   tokenizer plays an important role in the per-language downstream task performance, which \citet{Clark2021Canine, Xue2021ByT5, Tay2021Charformer} take to the extreme by proposing tokenizer-free approaches.  

To extend a monolingual LM to other languages, \citet{artetxe-etal-2020-cross} train a new embedding layer with a corresponding target-language tokenizer, while freezing the pre-trained transformer weights. \citet{Ke2020} extend a monolingual model to new languages using bilingual corpora. \citet{wang-etal-2020-extending, chau-etal-2020-parsing} extend the vocabulary of multilingual models with a small number of target-language tokens, to improve the performance in the target language. \citet{muller-etal-2021-unseen} propose a transliteration based approach, \citet{vernikos-popescu-belis-2021-subword-mapping} propose subword mappings, and   \citet{pfeiffer-etal-2020-mad,pfeiffer2020unks,Vidoni2020OrthogonalLA,Ansell2021MADG} propose adapter-based approaches to extend multilingual models to unseen languages.

While these approaches achieve considerable performance gains over unseen languages, they are outperformed by  standard full fine-tuning methods for  seen languages. 
One can further argue that, as the pre-trained models have already been cursed by multilinguality, the adapter-based approaches build upon sub-optimal parameter initializations.\footnote{We investigate this claim further in \S\ref{sec:xmod_ada}.} 
In our work, we consequently aim to \textbf{1)} modularize the model from the start to prepare the model to be \textbf{2)} extendable to new languages post-hoc.

\section{Proposed approach}

We propose \textsc{X-Mod}, a modular multilingual architecture that combines shared and language-specific parameters. In contrast to prior work,  
we pre-train modular models from the get-go. Our models can be extended to new languages after pre-training, and used for cross-lingual transfer learning in downstream tasks.

\vspace{0.2em}
\noindent\textbf{Architecture.} As illustrated in Figure~\ref{fig:modular_architecture}, we extend the transformer-based architecture from mBERT \cite{Devlin2019bert} and XLM-R \citep{Conneau2020xlm-r} by incorporating language-specific modules---bottleneck feed-forward layers---at every transformer layer. We learn a separate module for each language, whereas the attention and feed-forward components are shared. While the total number of parameters of the model grows linearly with the number of languages, the training and inference cost does not increase (as measured in FLOPs), as only the module in the relevant language is used for each input. Inspired by the adapter\footnote{The term `adapter' refers to newly introduced layers within a pre-trained (frozen) model. These layers \textit{adapt} the representations of the pre-trained mode; we train these modular components together with the transformer weights, and therefore refer to them as modules. } architecture of \citet{pfeiffer-etal-2021-adapterfusion} we place our `modules' after the LayerNorm of the feed-forward transformer block, and the residual connection is placed after the LayerNorm;\footnote{We find that the residual connection proposed by \citet{pfeiffer-etal-2021-adapterfusion} results in training instabilities when trained together with the transformer weights.} the LayerNorm before and after the modular component is shared.\footnote{Preliminary results showed that sharing the LayerNorm results in better cross-lingual transfer performance.}  

\vspace{0.2em}
\noindent\textbf{Pre-training procedure.} Similar to \citet{Conneau2020xlm-r}, we pre-train our model on MLM on combined monolingual corpora in multiple languages. Examples of each language are passed through the shared embedding matrix as well as the multi-head attention and feed-forward components at each layer. As each layer contains a language-specific modular component, the examples are routed through the respective designated modular bottleneck layer. Given that each example only requires access to a single module, modules can be efficiently stored on only a subset of GPUs in distributed training. 

\vspace{0.2em}
\noindent\textbf{Extending to new languages.} The modular design of our model allows us to  extend it to new languages after pre-training. To that end, we learn new embeddings and adapter modules for the target language through MLM, while the rest of the components are frozen.\footnote{Following \citet{artetxe-etal-2020-cross} we  train positional embeddings.} Consequently, we are able to extend the model to a new language by learning a small number of new parameters, without affecting performance in the set of pre-trained languages. Following \citet{pfeiffer2020unks}, we learn a new subword vocabulary for the added languages, and initialize the embeddings of lexically overlapping tokens from the original embedding matrix. 

\vspace{0.2em}
\noindent\textbf{Fine-tuning on downstream tasks.} To transfer the  models to cross-lingual downstream tasks, we fine-tune the shared weights only on the source language data, while keeping the modular components and the embedding layer frozen. We follow the standard fine-tuning procedure of adding a prediction head on top of the CLS token. We then replace the source language modules (as well as embedding layer for \textit{added} languages) with the target language parameters, passing the text of the target language through the model.\footnote{We initially also experimented with stacking adapters on top of the language modules similar to \citet{pfeiffer-etal-2020-mad, pfeiffer2020unks}. While this approach is considerably more parameter efficient, we find that fine-tuning all shared weights slightly outperformed the adapter-based approach.}

\begin{table*}[t!]
\begin{center}
\begin{small}
\addtolength{\tabcolsep}{-0.5pt}
\begin{tabular}{ccl}
\toprule
\multirow{6}{*}{\shortstack{Pre-trained \\ languages}}
& 13-LANGS &
\underline{\textbf{\textit{en}}}, \underline{\textbf{\textit{ar}}}, \underline{\textbf{fr}}, \underline{\textbf{\textit{hi}}}, \underline{ko}, \underline{\textbf{\textit{ru}}}, \underline{\textbf{\textit{th}}}, \underline{\textbf{\textit{vi}}}, \underline{ta}, \underline{id}, \underline{fi}, \textbf{\underline{sw}}, \underline{ka} \\
\cmidrule{2-3}
& 30-LANGS & 13-LANGS +
cs, eu, hr, hu, hy, it, lt, ml, mn, ms, pl, ro, si, sk, sq, sv, tl \\
\cmidrule{2-3}
& \multirow{2}{*}{60-LANGS} & 30-LANGS + af,  am, be, bn, ca, cy, da, eo, et, fa, ga, gl, gu, ha, is, ku, la, lv, mk, ne, nl, no, ps, \\ 
& &   pt, sa, sd, sl, so, sr, te \\
\cmidrule{2-3}
& 75-LANGS & 60-LANGS + as, br, bs, fy, gd,  jv, kn, mg, mr,  om, or, pa,  su, xh, yi,   \\
\midrule
\multicolumn{2}{c}{Added languages} &

\underline{\textbf{bg}},
\underline{\textbf{\textit{de}}},
\underline{\textbf{\textit{el}}},
\underline{\textbf{\textit{es}}},
\underline{\textbf{\textit{tr}}},
\underline{\textbf{ur}},
\underline{\textbf{\textit{zh}}},
\\
\bottomrule
\end{tabular}
\end{small}
\end{center}
\vspace{-0.7em}
\caption{\textbf{Selection of languages.} We pre-train different models on 4 sets of languages, and further extend them to a set of held-out languages post-hoc. We evaluate on XNLI (languages in \textbf{bold}), NER (\underline{underlined} languages) and XQuAD/MLQA (languages in \textit{italic}). For more details about the   language  selection, see  Appendix \ref{app:language_selection}.}
\label{tab:languagelist}
\vspace{-0.5em}
\end{table*}

\section{Experimental design}

We  detail the baseline and models (\S\ref{sec:model_variants}), and their training  (\S\ref{sec:training_details}) and evaluation settings (\S\ref{sec:evaluation}).

\subsection{Model variants}
\label{sec:model_variants}
 
We pre-train  
separate models for all combinations along the following axes: 

\vspace{0.2em}
\noindent\textbf{\textsc{X-Mod} vs. \textsc{shared}.} 
To evaluate the effectiveness of our \textsc{X-Mod} model, we aim to compare ourselves to a conventional non-modular architecture. However, simply removing the modular component would be unfair, as the number of FLOPs and trainable parameters per language would not be the same---both in terms of pre-training, as well as fine-tuning. Consequently, for our baseline model---where all parameters should be \textit{fully} shared between all languages---we include a single bottleneck layer right after the Feed-Forward component. Effectively, this is the same architecture as our \textsc{X-Mod} model, just with a single module that is shared by all languages. We refer to this as the \textsc{shared} model throughout this paper.\footnote{Extending the \textbf{total} number of shared parameters would be unfair, 
as \textsc{X-Mod} and  \textsc{shared}  would not have the same FLOPs nor the same number of trainable parameters when fine-tuning. } 
To extend the \textsc{shared} model to unseen languages, we follow \citet{artetxe-etal-2020-cross} and only learn a new embedding layer, freezing the transformer parameters. To fine-tune the \textsc{shared} model on a downstream task, we freeze the embedding layer, as well as the (single) module, thereby fine-tuning an equal amount of parameters on the downstream task as the \textsc{X-Mod} model.\footnote{Adapter-based approach such as MAD-X \cite{pfeiffer-etal-2020-mad} would be an alternative. However, this would require training on languages twice---once during pre-training, and once when adding adapters---which is not directly comparable to \textsc{X-Mod}. Nonetheless, we report results in \S\ref{sec:xmod_ada}.} 

\vspace{0.2em}
\noindent\textbf{13 vs. 30 vs. 60 vs. 75 languages.} So as to understand how each approach is affected by the curse of multilinguality, we pre-train the \textsc{X-Mod} and \textsc{shared} models on 4 increasing sets of languages. We start with an \textit{initial} set of 13 typologically diverse  languages that we evaluate on, and add additional languages for larger sets of 30, 60, and 75 languages. In addition, we keep a set of 7 held-out languages that we extend the pre-trained models to. Table \ref{tab:languagelist} lists the specific languages in each group. The selection and split of \textit{initial} as well as \textit{added} languages is motivated by typological and geographical diversity, as well as the availability of downstream task evaluation data.

\vspace{1em}
\noindent\textbf{Controlling for total vs. per-language updates.} \citet{Conneau2020xlm-r} investigated the effect of adding more languages during pre-training, while training on an equal number of update steps. However, increasing the number of languages while keeping the number of updates constant results in the model seeing less data in each individual language. As such, it remains unclear if the curse of multilinguality happens because of negative interference, or simply because the number of updates for each specific language is smaller.   
So as to understand this, we compare (1) training on an equal number of \textit{update steps} and (2) training on an equal number of \textit{seen examples} per language. 
We start with the set of 13  languages (Table~\ref{tab:languagelist}) and train the respective models for 125k update steps. When adding more languages, we compare (1) training models on each set of languages for 125k update steps, and (2) increasing the number of update steps such that the models are trained on the same number of examples in each of the initial 13 languages. For the latter, this amounts to training for 195k, 265k and 269k update steps, 
respectively.

\subsection{Training details}
\label{sec:training_details}

\noindent\textbf{Data and hyperparameters.}  
We sample languages with $\alpha=0.7$ 
and train our models with a batch size of 2048 across 64 V100 GPUs on the CC100 dataset \cite{Conneau2020xlm-r} using  fairseq  \cite{ott-etal-2019-fairseq}.  
All our models extend the \textit{base} transformer architecture, with 12 layers and 768 dimensions.  Modules are implemented with a bottleneck size of 384. The shared transformer weights account for 270M parameters, whereas each individual module accounts for 7M parameters. We train our models with a linear learning rate decay peaking at $7\mathrm{e}{-4}$ during pre-training and  $1\mathrm{e}{-4}$ when adding languages.

\vspace{0.2em}
\noindent\textbf{Vocabulary.} 
As we aim to identify the impact of \textit{modularity} on the curse of multilinguality, we  control for consistent tokenization across the different axes. We therefore tokenize using the XLM-R vocabulary for all our pre-training experiments.\footnote{\citet{rust-etal-2021-good} have previously demonstrated the impact  of the multilingual tokenizer on the downstream task performance: languages underrepresented in the sub-word vocabulary exhibit considerable performance drops when compared to vocabularies dedicated to the respective language. } However, for  languages added post-hoc, we learn a \textit{new} SentencePiece tokenizer for each of the target language,\footnote{We train the new tokenizers for a vocabulary size of 30k. } as the languages potentially use scripts unseen by the original tokenizer.

\subsection{Evaluation}
\label{sec:evaluation}

We conduct experiments on NLI, NER, and QA. In all cases, we fine-tune the model on English and measure the zero-shot transfer performance in other languages. For NLI we train on MultiNLI \cite{williams-etal-2018-broad} and evaluate on XNLI \cite{conneau-etal-2018-xnli}. For QA, we train on SQuAD \cite{Rajpurkar2016squad} and evaluate on XQuAD \cite{artetxe-etal-2020-cross} and MLQA \cite{lewis-etal-2020-mlqa}. For NER, we use WikiANN \cite{Pan2017wikiann, Rahimi2019massively}.  
We  
experiment
with learning rates $1\mathrm{e}{-4}$, $3\mathrm{e}{-4}$, and $5\mathrm{e}{-4}$ and train for 3 or 5 epochs for QA and 5 or 10 epochs for NER and NLI. For NER and NLI we take the  hyperparameter setting performing best on the development sets, averaged across the pre-trained languages (Table~\ref{tab:languagelist}). For SQuAD we take the best performing checkpoint evaluated on the English development set, and report the cross-lingual test set results.\footnote{In contrast to NER and NLI, the cross-lingual evaluation benchmarks of SQuAD do not provide a development set for each target language on the basis of which the best checkpoint can be selected. Consequently, we select the checkpoint based on the best performance on the English development set. }  
All results are averaged  across 5  random seed runs.

\begin{figure*}[t!]
    \centering
    \begin{subfigure}[!t]{\linewidth}
        \centering
        \includegraphics[width=0.985\linewidth]{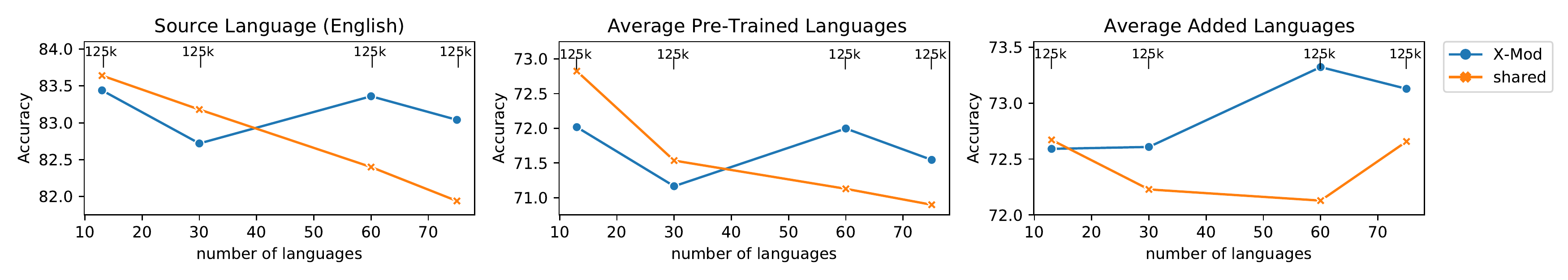}
    \end{subfigure}
        \begin{subfigure}[!t]{\linewidth}
        \centering
        \includegraphics[width=\linewidth]{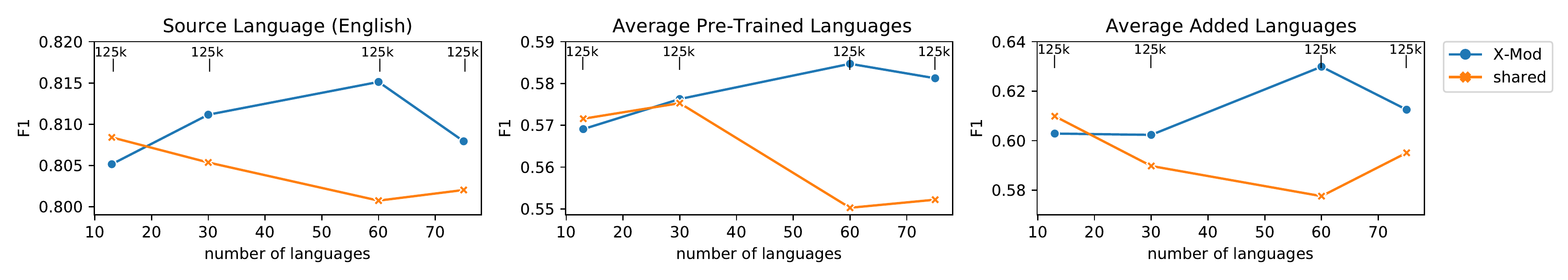}
        \caption{ All models are trained for 125k update steps. Models trained on \textbf{more languages} have seen \textbf{less examples} in each  language.}
        \label{fig:task_125k}
    \end{subfigure}

    \begin{subfigure}[!t]{\linewidth}
        \centering
        \includegraphics[width=0.985\linewidth]{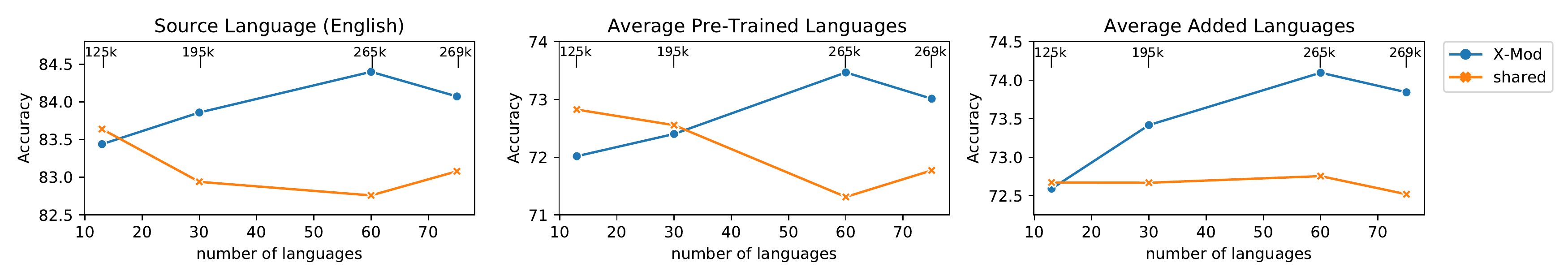}
    \end{subfigure}
        \begin{subfigure}[!t]{\linewidth}
        \centering
        \includegraphics[width=\linewidth]{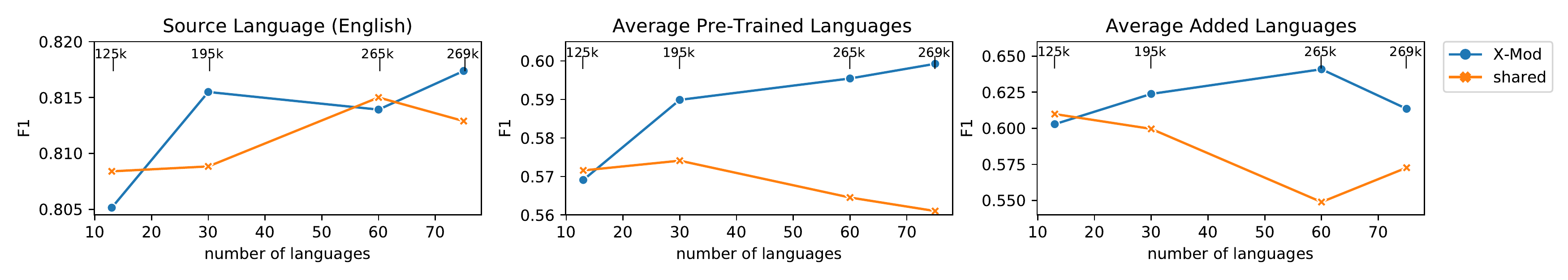}
        \caption{ Models trained on more languages are trained longer. All models have seen the \textbf{same amount of examples} in each  language.}
        \label{fig:task_same}
    \end{subfigure}
    \caption{Test set results on XNLI (top) and NER (bottom) for models trained on different numbers of languages.  
    \textit{Source Language (English)} only includes scores of the source language. \textit{Average Pre-Trained Languages} includes all evaluation languages that the model was pre-trained on. \textit{Average Added Languages} includes all languages that were added to the model after pre-training. Scores are averaged across all languages and random seeds. 
    }
\label{fig:maintaining_update_steps}
\end{figure*}

\section{Results and discussion}

\begin{table*}[t!]
\begin{center}
\def\arraystretch{0.87}
\resizebox{0.99\textwidth}{!}{%
\addtolength{\tabcolsep}{-0.5pt}
\begin{tabular}{clc|cccccccccccc|c}
\toprule
&& en & ar & fr & hi & ko & ru & th & vi & ta & id & fi & sw & ka & avg \\
\midrule
\multirow{2}{*}{NER}
& \textsc{X-Mod} & 81.4 & \bf 78.9 &  \bf 77.2 &  \bf 70.1 &  \bf 53.0 &  \bf 59.1 & 2.8 & \bf  66.2 & 51.1 & 50.5 &  \bf 78.6 & \bf  73.4 &  \bf 67.3 &  \bf 62.8 \\
& \textsc{shared}  & \bf 81.5 & 74.1 & 74.7 & 64.4 & 46.0 & 58.3 & \bf  4.0 & 63.7 & \bf  52.5 & \bf  51.5 & 74.4 & 57.2 & 61.5 & 58.8 \\
\midrule
\multirow{2}{*}{XNLI}
& \textsc{X-Mod} &  \bf 84.4 &  \bf 71.2 &  \bf 77.6 & \bf  68.3 & - &  \bf 74.1 & \bf  71.7 &  \bf 73.4 & - & - & - &  \bf 66.9 & - &  \bf 73.5 \\
& \textsc{shared}  & 82.8 & 69.2 & 75.6 & 66.6 & - & 73.2 & 68.5 & 72.5 & - & - & - & 62.1 & - & 72.5 \\
\midrule
\multirow{2}{*}{XQuAD}
& \textsc{X-Mod} & \bf 85.1 & \bf 68.1 & - & \bf 67.5 & - & \bf 75.0 & \bf 66.3 & \bf 74.9 & - & - & - & - & - & \bf 72.8 \\
& \textsc{shared}  & 83.8 & 64.6 & - & 65.8 & - & 72.7 & 63.0 & 72.6 & - & - & - & - & - & 70.4 \\
\midrule
\multirow{2}{*}{ MLQA}
& \textsc{X-Mod} & \bf 80.1 & \bf 58.6 & - & \bf 60.7 & - & - & - & \bf 67.5 & - & - & - & - & - & \bf 66.7 \\
& \textsc{shared}  & 79.6 & 53.6 & - & 58.7 & - & - & - & 64.9 & - & - & - & - & - & 64.2 \\

\bottomrule
\end{tabular}
}
\end{center}
\caption{Pre-trained language results for the modular and shared model variants, pre-trained on the set of 60 languages for 265k update steps. For NER and MLQA we report \textit{F$_1$}, for XNLI \textit{accuracy} scores. Scores are averaged across all 5 random seeds of the best hyperparameter setting, evaluated  on the development set.  
}
\label{tab:unified-results}
\end{table*}

We present results for pre-trained languages in \S\ref{sec:pre-trained_langs} and added languages in \S\ref{sec:added_langs}.

\subsection{Pre-trained languages}
\label{sec:pre-trained_langs} 
 
In Figure~\ref{fig:maintaining_update_steps} we plot downstream task results of models pre-trained on different amounts of languages. 
Table~\ref{tab:unified-results} reports the individual language performance for the  models trained on 60 languages.

\vspace{0.2em}
\noindent\textbf{The Curse of Multilinguality.}
\citet{Conneau2020xlm-r} showed that  multilingual LMs  trained on \textit{increasing} amounts of languages, while \textit{maintaining} the  number of update steps,  exhibit drops in downstream task XNLI performance. We reproduce these results, both in terms of language modelling perplexity (Figure~\ref{fig:Mean_Perplexity}),\footnote{For  per-language perplexity see Appendix \ref{app:additional_results}.}  as well as downstream task performance on XNLI \textit{and} NER (Figure~\ref{fig:task_125k}). 
We further find that the curse of multilinguality does not \textit{only} happen \textit{because} the total number of update steps per language decreases, but \textit{also} when all \textsc{shared} models are trained on the \textit{same} number of examples per language (Figure~\ref{fig:task_same}). This confirms that fully shared architectures  suffer from negative interference. 

\vspace{0.2em}
\noindent\textbf{Lifting the Curse.} 
While for the \textsc{shared} model we witness 
 negative interference between languages
in terms of perplexity, the \textsc{X-Mod} model is able to \textit{maintain} performance, and even improves for a subset of languages. We observe similar patterns in the downstream task performance:
In both our experimental setups---(1) we control for the number of update steps (Figure~\ref{fig:task_125k}); (2) we control for the number of per-language seen examples (Figure~\ref{fig:task_same})---our 
 \textsc{X-Mod} model---in contrast to the \textsc{shared} model---is able to maintain, or even outperform model variants trained on less languages. These results demonstrate that the  added per-language capacity is sufficient for the model to adequately represent all languages.

Surprisingly, \textsc{X-Mod} not only maintains performance, but actually slightly improves while we  increase the number of  languages we pre-train on. This is even the case for settings where the  model sees \textit{less} examples in the target language. This suggests that increasing the language diversity can have a positive impact on the model's cross-lingual representation capability.

\vspace{0.2em}
\noindent\textbf{\textsc{X-Mod} \underline{vs} \textsc{shared}.}
Overall, the \textsc{X-Mod} model pre-trained on 60 languages  achieves the best cross-lingual performance.\footnote{We find that the \textsc{X-Mod} model trained on 75 languages is  less stable than the versions trained on less languages. We think that this can be attributed to the 15 added languages being extremely low resource---we only train for an additional 4k update steps---resulting in the respective randomly initialized modules being updated very infrequently.
 This variance could potentially be mitigated by  training for longer.} 
 Our results on XNLI, NER, MLQA, and XQuAD in Table~\ref{tab:unified-results} demonstrate consistent performance gains  over the \textsc{shared} model for every task and across (almost) all high- as well as low-resource languages.

\begin{table}[t!]
\begin{center}
\def\arraystretch{0.87}
\resizebox{1.0\columnwidth}{!}{%
\addtolength{\tabcolsep}{-0.5pt}
\begin{tabular}{clccccccc|c}
\toprule
&& bg & de & el & es & tr & ur & zh& avg \\

\midrule
\multirow{2}{*}{NER}
& \textsc{X-Mod} & \bf 77.6 & \bf 75.1 & \bf 75.2 & \bf 71.9 & \bf 72.6 & \bf 54.7 & \bf 21.6 & \bf 64.1 \\
& \textsc{shared}  & 74.9 & 66.3 & 69.6 & 49.1 & 64.8 & 50.4 & 9.2 & 54.9 \\
\midrule
\multirow{2}{*}{XNLI}
& \textsc{X-Mod} & \bf 77.4 & \bf 75.4 & \bf 76.2 & \bf 78.5 & \bf 72.4 & \bf 64.9 & \bf 73.8 & \bf 74.1 \\
& \textsc{shared}  & 76.3 & 74.1 & 74.9 & 77.3 & 71.0 & 64.3 & 71.4 & 72.8 \\
\midrule
\multirow{2}{*}{MLQA}
& \textsc{X-Mod} & -  & \bf 63.8  & -   & \bf 68.6 & -   & -   & \bf 61.7  & \bf  64.8 \\
& \textsc{shared}  & -  & 58.9  &  - & 66.7  & -  & -  & 56.5  & 60.7  \\

\bottomrule
\end{tabular}
}
\end{center}
\caption{Results for added languages, for models pre-trained on the set of 60 languages for 265k update steps. We report \textit{F$_1$} and \textit{accuracy} scores which are averaged across all 5 random seeds of the best hyperparameter setting on the development set.}
\label{tab:unified-results-added}
\end{table}

\subsection{Extending to unseen languages}
\label{sec:added_langs}

We further evaluate the cross-lingual performance of languages added in the second step; (1) on the architectural side---comparing the \textsc{shared} with the \textsc{X-Mod} modelling variant---and (2) by comparing the performance when \textit{pre-training} on the language, vs. when \textit{adding} the language post-hoc.

\vspace{0.2em}
\noindent\textbf{Modular vs Shared.}
We  evaluate if the additional per-language capacity improves the extendability  of the \textsc{X-Mod} model. On the right in Figure~\ref{fig:task_125k} we plot the results for added languages on XNLI (top) and NER (bottom). Similarly, we plot the results for the models where we control for the number of seen examples per target language in Figure~\ref{fig:task_same}. We find that the \textsc{X-Mod} model consistently outperforms the \textsc{shared} model, with a peak performance when pre-training on 60 languages, demonstrating that the language specific capacity is beneficial for adding new languages post-hoc. 
We report results for the 60 language versions in Table~\ref{tab:unified-results-added}, demonstrating the consistent advantage of the \textsc{X-Mod}  over the \textsc{shared} model.

\vspace{0.2em}
\noindent\textbf{Pre-training vs Adding Languages.} 
To evaluate if there is a measurable difference on downstream performance for languages that we \textit{pre-train} on vs. those we \textit{add post-hoc}, we train 2 models on \textit{different} initial sets of languages, adding the respectively missing ones in the second step.  
So as to understand if the typological similarity of languages has impact on the downstream task performance, we split the \textit{initial} and \textit{added} languages (Table~\ref{tab:languagelist}) of our previous experiments into two parts. The \textit{first} split consists of languages where the model was pre-trained on at least one language of the same language family (e.g. English vs. German). The \textit{second} split consists of languages that are part of a \textbf{unique} language family, i.e. the model was \textbf{not} pre-trained on a language of the same family  
(Table~\ref{tab:flipped-languagelist}). Consequently, we pre-train two models on two sets of languages, adding the respective other set  post-hoc.\footnote{In  previous experiments, the modular model trained on 60 languages achieved  the best performance. Therefore, the models in these experiments are also trained on 60 languages. Both models are trained on the same additional languages, i.e. the 60-LANGS of Table~\ref{tab:languagelist}, where only the  13-LANGS differ.}

\begin{figure}[t]
    \centering
        \includegraphics[width=\columnwidth]{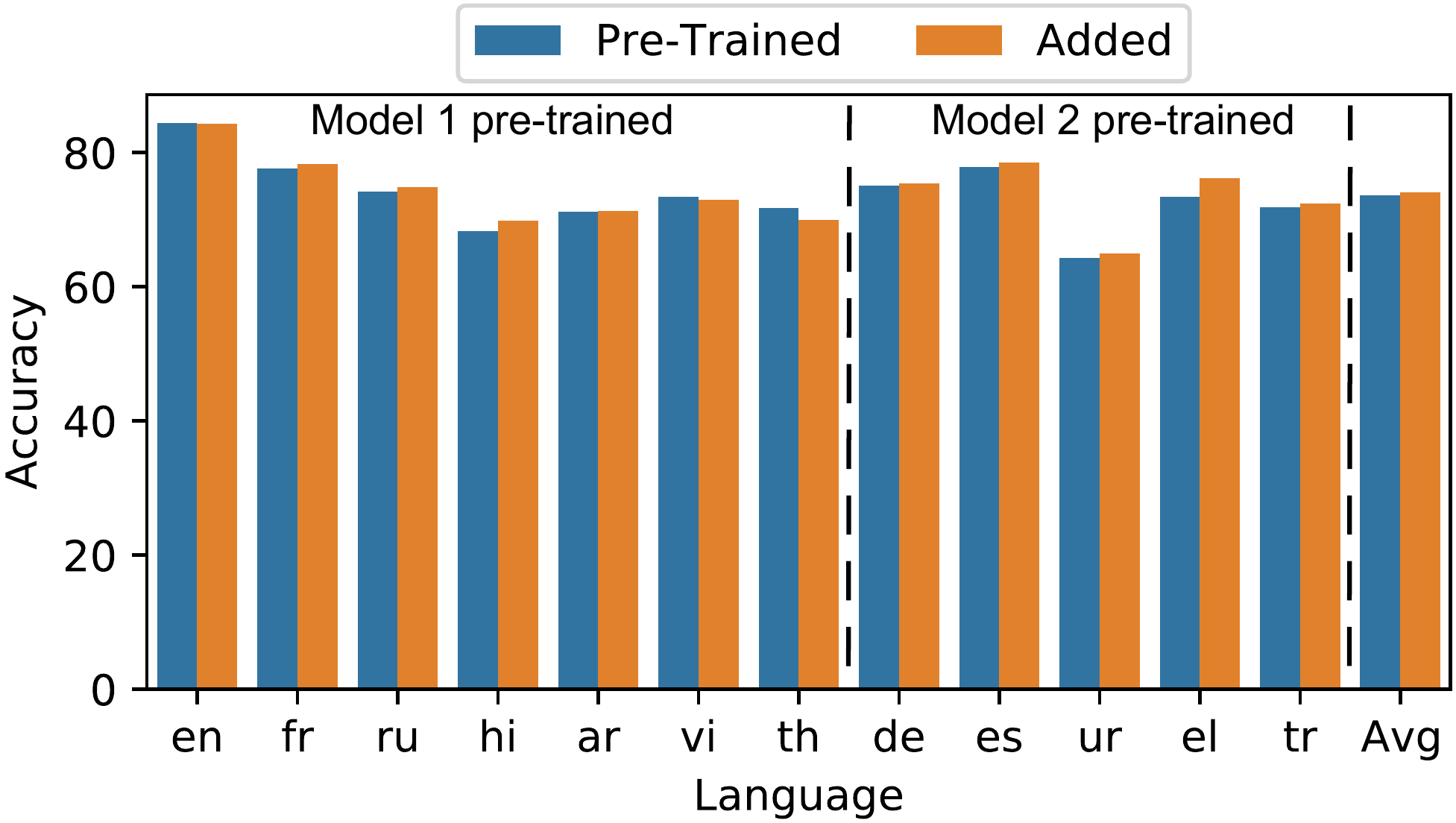}
    \caption{XNLI test set accuracy of \textsc{X-Mod} models pre-trained on different  languages in comparison to those added post-hoc (Table~\ref{tab:flipped-languagelist}). }
\label{fig:NLI_added_vs_pre-trained}
\end{figure}

\begin{table}[t]
    \centering
\resizebox{1.0\columnwidth}{!}{%
\begin{tabular}{llllccc}
\toprule

Language   & iso & Family         & Script   & Model 1 & Model 2 \\
\midrule
\rowcolor{lightgray}\textcolor{black}{ English}    & \textcolor{black}{en}  & \textcolor{black}{IE: Germanic}   & \textcolor{black}{Latin}    & \textcolor{black}{pre-train} & \textcolor{black}{add} \\
 German     & de  & IE: Germanic   & Latin    &  \textcolor{black}{add} & \textcolor{black}{pre-train}  \\
\rowcolor{lightgray} \textcolor{black}{French}     & \textcolor{black}{fr}  & \textcolor{black}{IE: Romance}    & \textcolor{black}{Latin}  &  \textcolor{black}{pre-train} & \textcolor{black}{add}  \\
Spanish    & es  & IE: Romance    & Latin  &  \textcolor{black}{add} & \textcolor{black}{pre-train}  \\
\rowcolor{lightgray} \textcolor{black}{Russian}    & \textcolor{black}{ru}  & \textcolor{black}{IE: Slavic}     & \textcolor{black}{Cyrillic}  &  \textcolor{black}{pre-train} & \textcolor{black}{add}\\
Ukranian   & uk  & IE: Slavic     & Cyrillic   &  \textcolor{black}{add} & \textcolor{black}{pre-train}\\
\rowcolor{lightgray} \textcolor{black}{Hindi}    & \textcolor{black}{hi}  & \textcolor{black}{IE: Iranian}     & 	\textcolor{black}{Devanagari}   &  \textcolor{black}{pre-train} & \textcolor{black}{add}\\
Urdu   & ur  & IE: Iranian     & Arabic  &  \textcolor{black}{add} & \textcolor{black}{pre-train}\\
\rowcolor{lightgray} \textcolor{black}{Arabic}     & \textcolor{black}{ar}  & \textcolor{black}{Afro-Asiatic}   & \textcolor{black}{Arabic}    & \textcolor{black}{pre-train} & \textcolor{black}{add}\\
Hebrew     & he  & Afro-Asiatic   & Hebrew     &  \textcolor{black}{add} & \textcolor{black}{pre-train}\\
\midrule
\rowcolor{lightgray} \textcolor{black}{Vietnamese} & \textcolor{black}{vi}  & \textcolor{black}{Austro-Asiatic} & \textcolor{black}{Latin}      & \textcolor{black}{pre-train} & \textcolor{black}{add}\\
\rowcolor{lightgray} \textcolor{black}{Thai}       & \textcolor{black}{th}  & \textcolor{black}{Kra-Dai}        & \textcolor{black}{Thai}      &  \textcolor{black}{pre-train} & \textcolor{black}{add} \\
\rowcolor{lightgray} \textcolor{black}{Korean}     & \textcolor{black}{ko}  & \textcolor{black}{Koreanic}       & \textcolor{black}{Korean}  &  \textcolor{black}{pre-train} & \textcolor{black}{add}\\
Japanese   & ja  & Japonic        & Japanese  &  \textcolor{black}{add} & \textcolor{black}{pre-train}\\
Greek      & el  & IE: Hellenic   & Greek    &  \textcolor{black}{add} & \textcolor{black}{pre-train} \\
Turkish    & tr  & Turkic         & Latin     &  \textcolor{black}{add} & \textcolor{black}{pre-train} \\ 

\bottomrule
\end{tabular}
 
}
    \caption{Selection of 2 sets of languages that we either pre-train on, or add post-hoc. The last 6 languages in the list are part of language families which are \textit{unique} in the total list of languages we pre-train on (Table~\ref{tab:languagelist}), i.e. none of our models was pre-trained on a language of the same family. 
    }
    \label{tab:flipped-languagelist}
\end{table}

Our XNLI results (Figure~\ref{fig:NLI_added_vs_pre-trained}) demonstrate that the per-language performance is on par when pre-training vs. when adding the language post-hoc.\footnote{The models have seen an equal amount of examples in the respective languages in each case.}  
We also find that the  family does not have a measurable effect on the performance of the language. 
Our results therefore suggest that it is sufficient to train \textsc{X-Mod} on only a subset of languages for which sufficient pre-training data exists. Essentially, \textsc{X-Mod} has the potential to cover all languages of the world, as the model has the capability to be adapted to new languages post-hoc.

\section{Further analysis}

We further analyze the impact of the number of update steps on  \textsc{X-Mod}  (\S\ref{sec:modularity_kicking_in}) and compare our method to adapter-based approaches (\S\ref{sec:xmod_ada}).

\subsection{The importance of update steps}
\label{sec:modularity_kicking_in}

\begin{figure}[t!]
    \centering
        \includegraphics[width=\columnwidth,trim={0 0 0 0} , clip]{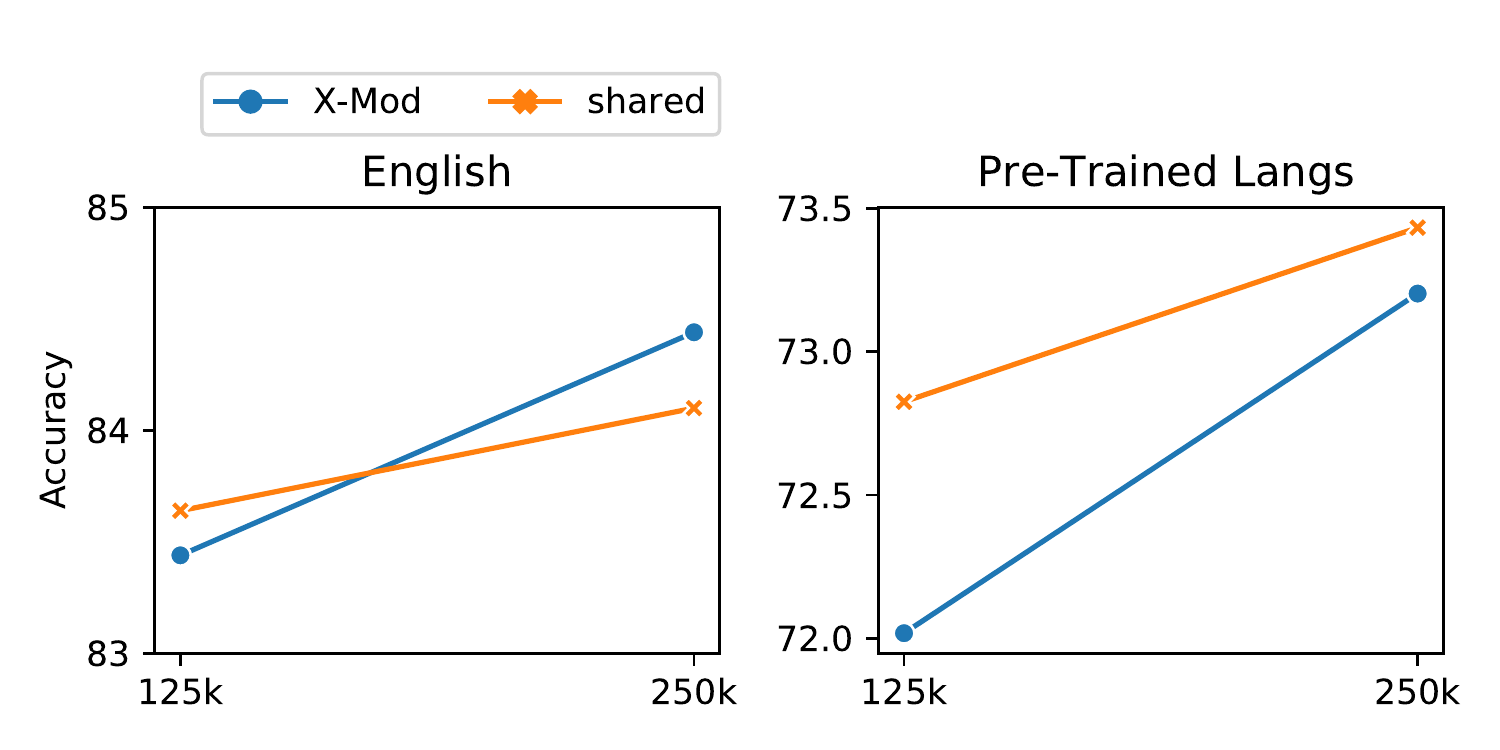}
    \caption{Results on XNLI when when pre-training on 13 languages for 125k and 250k update steps.}
\label{fig:NLI_13_longer}
\end{figure}

In Figure~\ref{fig:maintaining_update_steps}  we have witnessed a slight edge of the \textsc{shared} model over the \textsc{X-Mod} model, when training on only 13 languages and only training for 125k update steps. \citet{dufter-schutze-2020-identifying} found that it requires a large number of update steps for a model  pre-trained on multiple languages to become multilingual; with the added per-language capacity we hypothesize that update steps also play an important role for modular models.  
We compare the downstream task performance of models pre-trained on 13 languages, when training for 125k with  250k update steps in Figure~\ref{fig:NLI_13_longer}. When training for longer we find that the \textsc{X-Mod} model begins to outperforms the \textsc{shared} model in the source language, while almost closing the gap in the cross-lingual setting. This supports the hypothesis that the \textsc{X-Mod} model requires more  update steps   when training only on a small number of languages, in order for modularity to ``kick-in''.

\subsection{\textsc{X-Mod} vs. Adapters}
\label{sec:xmod_ada}

\begin{figure}[t!]
    \centering
        \includegraphics[width=\columnwidth,trim={0 0 0 0} , clip]{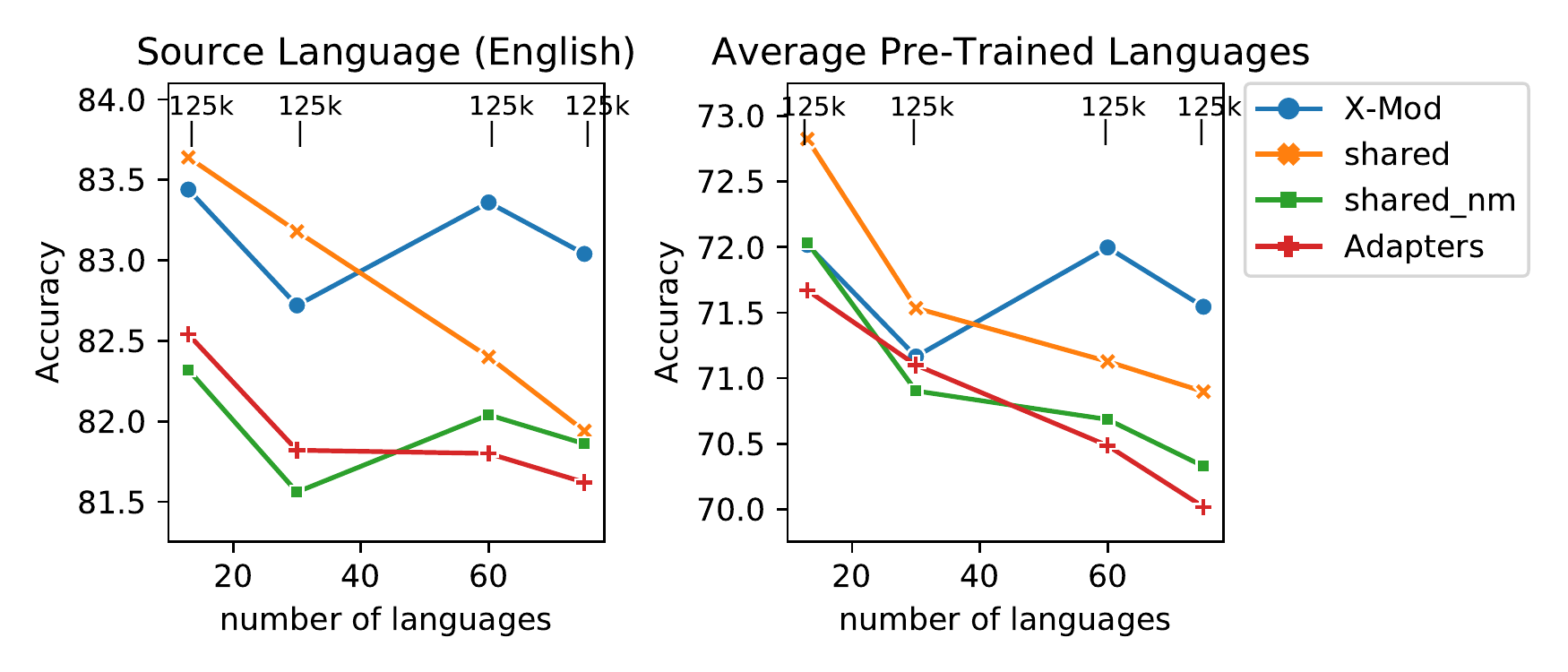}
    \caption{Comparison  on XNLI   of \textsc{X-Mod} and \textit{shared} models with an Adapter baseline, all  models are pre-trained for 125k update steps.}
\label{fig:NLI_adapters}
\end{figure}

As illustrated in Figure~\ref{fig:mod_ill}, from an architecture perspective \textsc{X-Mod} is similar to previously proposed multilingual Adapter-based methods \cite[\text{MAD-X;} ][]{pfeiffer-etal-2020-mad}. MAD-X utilizes a pre-trained massively multilingual transformer-based model and fine-tunes newly introduced adapter weights on languages the model has seen during pre-training, and ones the model has not been trained on. For a fair comparison in terms of \textit{seen examples} and \textit{number of update steps} we train a transformer model without module components (\textit{shared\_nm}) for 100k update steps on the respective languages (Table~\ref{tab:languagelist}). We subsequently train adapters on each of the target languages for another 25k update steps.\footnote{We follow \citet{pfeiffer-etal-2020-mad} and train adapter weights with a learning rate of 0.0001. While they have found that cross-lingual transfer performance of adapters converges at $\sim$20k update-steps, we would like to stress that our experimental setup is only \textbf{one} of multiple different valid versions. A more thorough investigation to find the optimal number of update steps for pre-training and subsequent adapter training is necessary, which was out of scope for this work.} We report results in comparison to \textsc{X-Mod} in Figure~\ref{fig:NLI_adapters}, here results for \textit{shared\_nm} are for a model that was trained for 125k update steps to instantiate a fair comparison.  

Our results demonstrate that the additional capacity of adapters added \textit{after} pre-training is not able to mitigate the curse of multilinguality which has already had a catastrophic impact on the shared transformer weights; the performance of the adapters strongly correlates with the performance of the corresponding fully shared model \textit{shared\_nm}. Consequently, adding language-specific capacity \textit{during} pre-training  is important, as the curse of multilinguality cannot be lifted post-hoc.

\section{Conclusions}

In this paper, we have evaluated the effectiveness of modular multilingual language modelling across multiple axes. We have demonstrated that by providing  additional per-language capacity, while maintaining the total number of trainable parameters per language, we are not only able to mitigate negative interference between languages, but additionally achieve  positive transfer. 
 
Our results suggest that it is sufficient to train our proposed \textsc{X-Mod} model only on a subset of  languages for which sufficient amounts of textual data is available.  
Unseen languages can be added post-hoc, with no measurable  drop in performance on XNLI.   
 By \textit{pre-training} the model in a modular fashion, we thus mitigate negative interference of idiosyncratic information, while simultaneously preparing the model to be extendable to unseen languages.

While in this work we have simulated language adding scenarios with a held out set of languages, in future work we aim to evaluate the performance  on truly low-resource languages such as MasakhaNER \cite{MasakhaNER} and AmericasNLI \cite{AmericasNLI}.
We further aim to evaluate the  cross-lingual transfer performance  from  typologically more diverse  source languages, besides English.

\section*{Acknowledgments}

We thank Samuel Broscheit for insightful feedback and suggestions on a draft of this paper, as well as the ARR reviewers and meta-reviewers for their valuable comments.

\bibliography{anthology,custom}
\bibliographystyle{acl_natbib}

\appendix


\section{Additional results} \label{app:additional_results}
We report MLQA and XQuAD results on pre-trained languages in Tables~\ref{tab:mlqa_pre-trained} and \ref{tab:xquad_pre-trained}, respectively, and MLQA results on added languages in Table~\ref{tab:mlqa_added}. Table~\ref{tab:ner_pre-trained} report NER results on more languages.

\begin{figure*}[t]
    \centering
        \includegraphics[width=\linewidth]{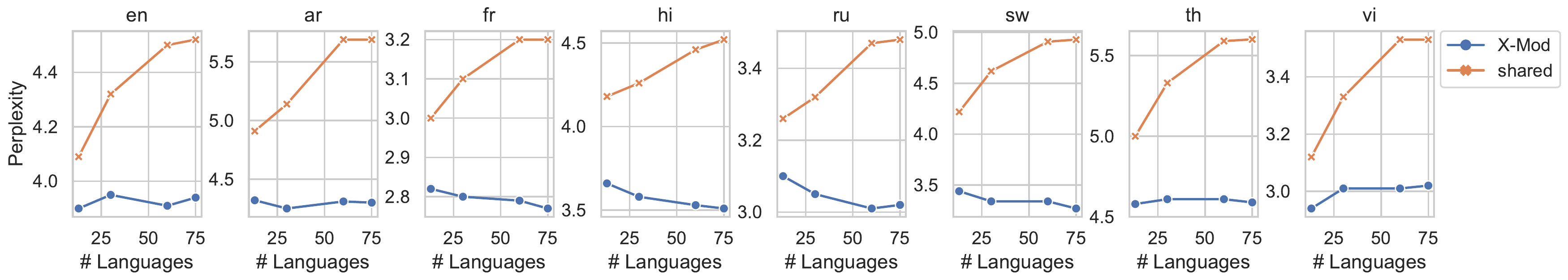}
    \caption{Perplexity  when training on more languages. Each model has seen the \textbf{same amount of examples} in each language. Lower perplexity indicates better performance. 
    }
\label{fig:perplexity_delta}
\end{figure*}

\begin{figure*}[t]
    \centering
    \begin{subfigure}[]{\linewidth}
        \centering
        \includegraphics[width=\linewidth]{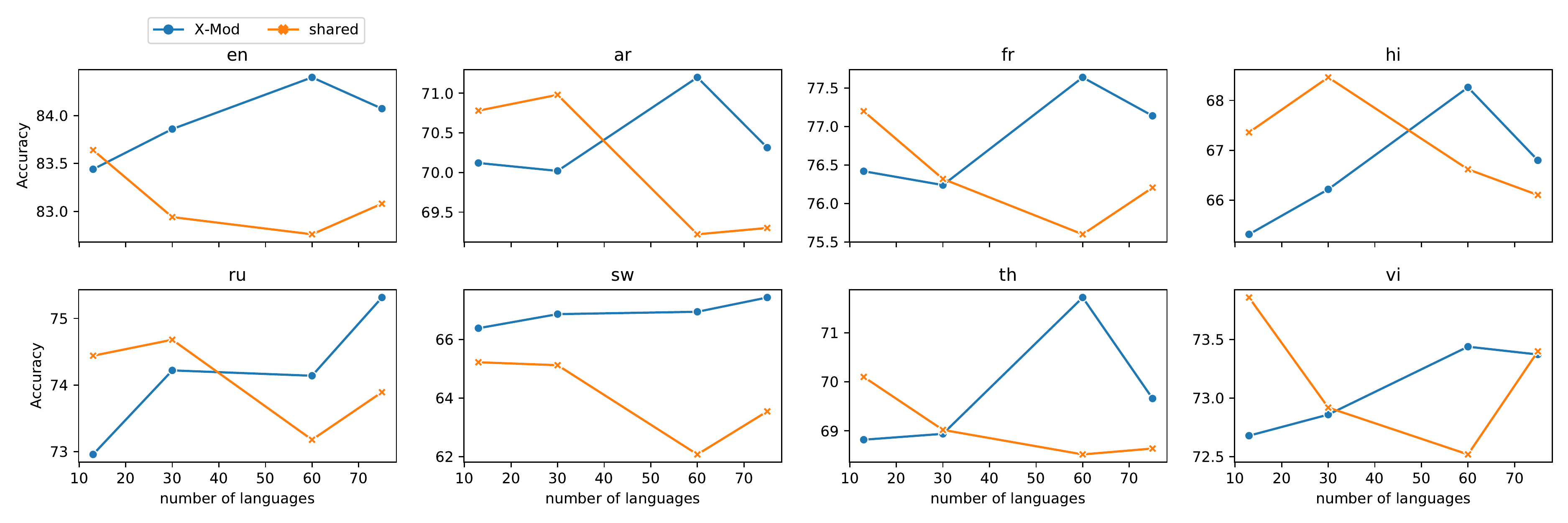}
        \caption{ Pre-Trained Languages} 
    \end{subfigure}
        \begin{subfigure}[]{\linewidth}
        \centering
        \includegraphics[width=\linewidth]{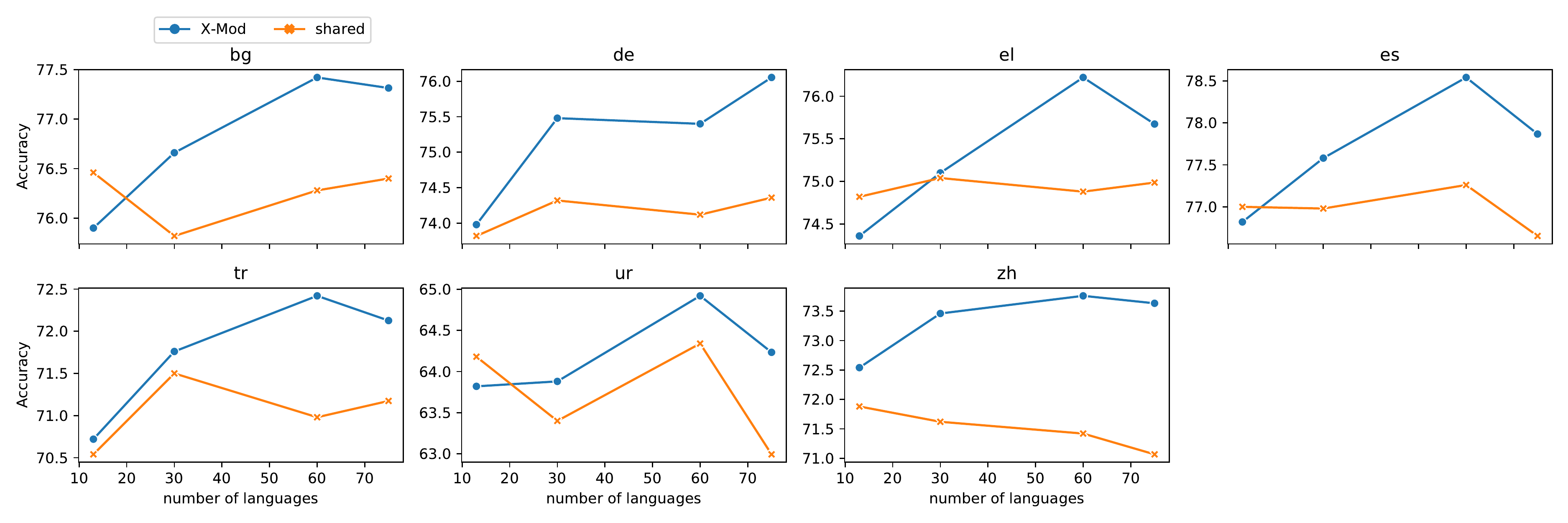}
        \caption{ Added Languages} 
    \end{subfigure}
  
    \caption{Testset results on \textbf{XNLI} of pre-trained (top) and added (bottom) languages trained on different numbers of languages. Models trained on more languages are trained for longer $\rightarrow$ all models have seen the \textbf{same amount of examples} in each individual language.  Scores are averaged across all  random seeds.  
    }
\label{fig:granular_NLI}
\end{figure*}

\begin{figure*}[b]
    \centering
    \begin{subfigure}[]{\linewidth}
        \centering
        \includegraphics[width=\linewidth]{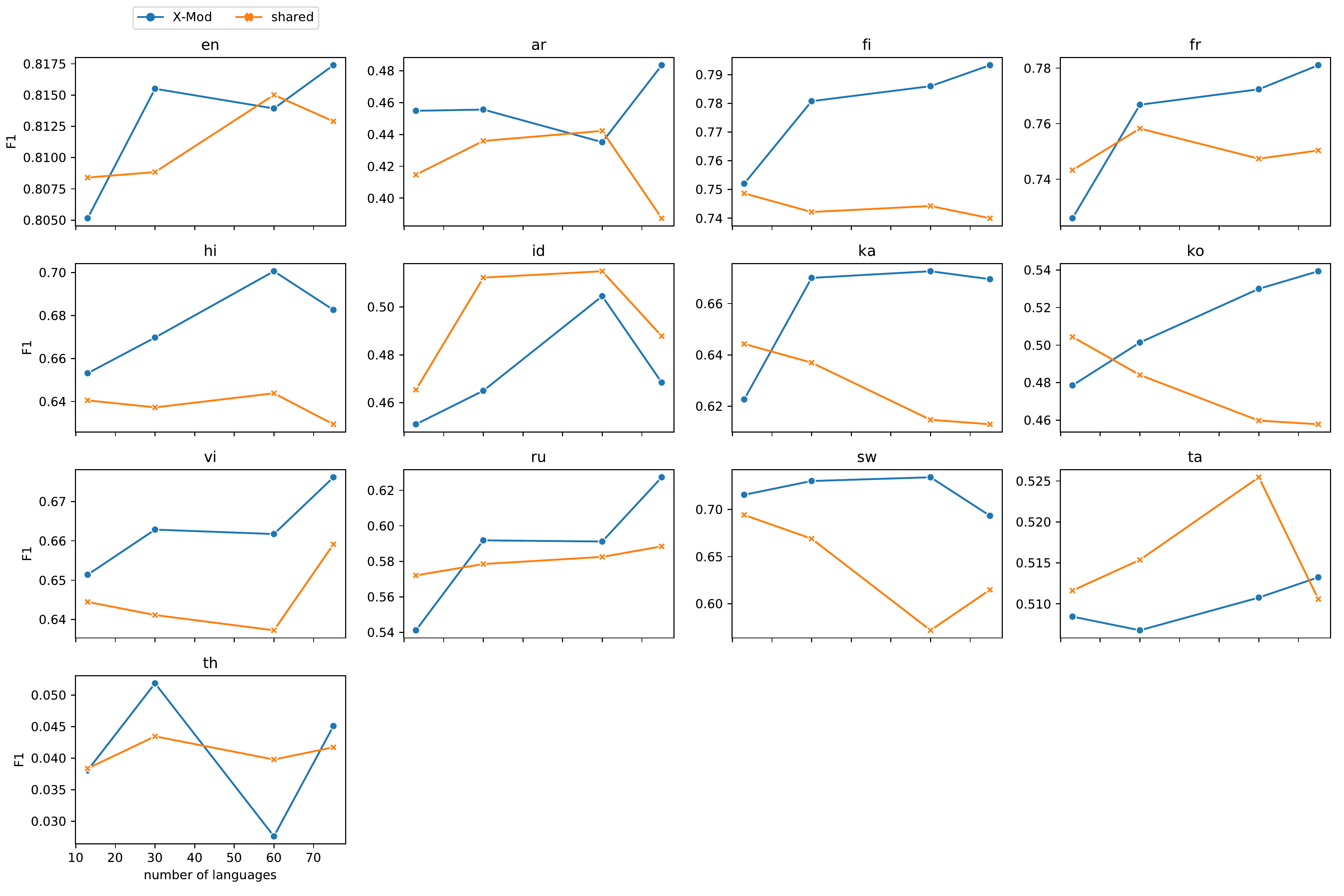}
        \caption{  Pre-Trained Languages} 
    \end{subfigure}
        \begin{subfigure}[]{\linewidth}
        \centering
        \includegraphics[width=\linewidth]{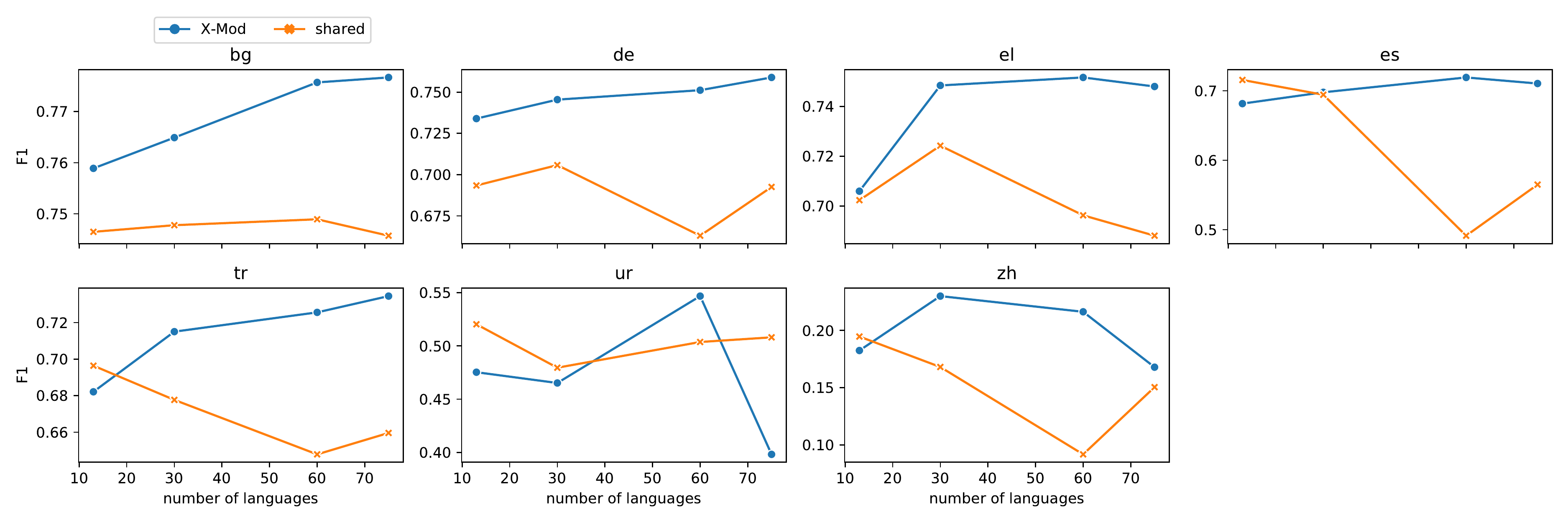}
        \caption{ Added Languages} 
    \end{subfigure}
  
    \caption{Testset results on \textbf{NER} of pre-trained (top) and added (bottom) languages trained on different numbers of languages. Models trained on more languages are trained for longer $\rightarrow$ all models have seen the \textbf{same amount of examples} in each individual language.  Scores are averaged across all  random seeds.  
    }
\label{fig:granular_NER}
\end{figure*}

Figures \ref{fig:perplexity_delta}, \ref{fig:granular_NLI} and \ref{fig:granular_NER} report per-language results as we increase the amount of languages on language modeling perplexity, XNLI and NER, respectively.

\section{Intermediate checkpoints} \label{app:intermediate_checkpoints} 
\begin{figure}[t!]
    \centering
        \includegraphics[width=\columnwidth]{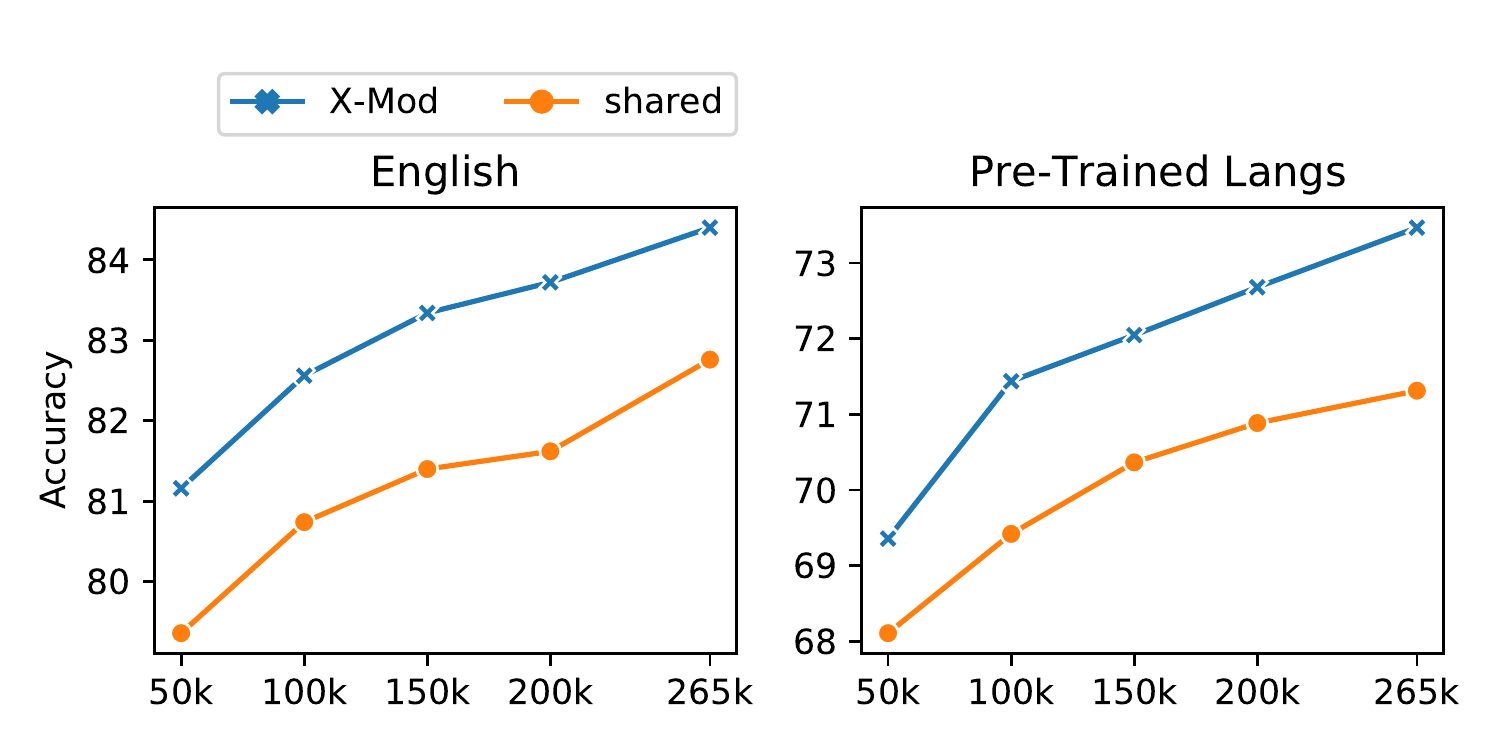}
        
    \caption{Results on XNLI  
    using intermediate checkpoints of the models trained on 60 languages. 
    }
\label{fig:NLI_intermediate_checkpoints}
\end{figure}

Our results in \S\ref{sec:modularity_kicking_in} suggest that, when the number of languages is small, \textsc{X-Mod} becomes more competitive with \textsc{shared} as the number of training steps increases. So as to understand if this behavior also holds for models covering more languages, we evaluate intermediate checkpoints for the 60-LANG model on XNLI. As shown in Figure~\ref{fig:NLI_intermediate_checkpoints}, we find that the \textsc{X-Mod} model continuously outperforms the \textsc{shared} model. This suggests that the \textsc{shared} model immediately suffers from negative interference between languages, while the added, language-specific components of the \textsc{X-Mod} model are able to mitigate the curse of multilinguality, resulting in considerable performance gains at all evaluated checkpoints.

\section{Language selection} \label{app:language_selection}
We provide more details about our selection of languages in Table~\ref{table:appendix_list_of_languages}.

\begin{table}[t!]
    \centering
        \def\arraystretch{0.87}
\resizebox{0.99\columnwidth }{!}{
\begin{tabular}{lrrrr|r}
\toprule
 &  \multicolumn{1}{c}{en}    &     \multicolumn{1}{c}{ar} &     \multicolumn{1}{c}{hi} &      \multicolumn{1}{c}{vi} &   \multicolumn{1}{|c}{\textit{avg}}\\
    & F$_1$ / EM & F$_1$ / EM & F$_1$ / EM & F$_1$ / EM & F$_1$ / EM \\
\midrule
\textsc{X-Mod} & \bf  80.1 / \bf 66.9    & \bf   58.6 / \bf 38.9   & \bf   60.7 / \bf 42.4   &     \bf     67.5 /  \bf 46.1  & \bf 66.7 / \bf  48.6   \\
\textsc{shared}  & 79.6 /  66.5   &  53.6 / 33.9    &  58.7 / 40.4    &  64.9 /    43.8       & 64.2  / 46.2   \\
\bottomrule
\end{tabular}
}
\caption{Average F$_1$ and Exact Match results for \textbf{pre-trained languages}, on the test set of \textbf{MLQA} for the \textsc{X-Mod} and \textsc{shared} model variants, pre-trained on the set of 60 languages for 265k update steps. \textbf{Bold} numbers indicate better performance for the respective language.}
\label{tab:mlqa_pre-trained}
\end{table}

\begin{table}[t!]
    \centering
        \def\arraystretch{0.87}
\resizebox{0.99\columnwidth }{!}{
\begin{tabular}{lrrrrrr|r}
\toprule
 &     \multicolumn{1}{c}{en} &     \multicolumn{1}{c}{ar} &     \multicolumn{1}{c}{hi} &     \multicolumn{1}{c}{ru} &     \multicolumn{1}{c}{th} &     \multicolumn{1}{c}{vi} &   \multicolumn{1}{|c}{\textit{avg}}\\
& F$_1$ / EM & F$_1$ / EM & F$_1$ / EM & F$_1$ / EM & F$_1$ / EM & F$_1$ / EM & F$_1$ / EM \\
\midrule
\textsc{X-Mod} &  \bf  85.1 /  \bf 73.4    & \bf  68.1 / \bf 52.4  & \bf 67.5 / \bf 50.3   & \bf 75.0 / \bf 57.8    & \bf  66.3 / \bf 52.6    & \bf  74.9 / \bf 54.6  & \bf 72.8 / \bf 56.9   \\
\textsc{shared}  &  83.8 / 72.1  &    64.6 / 48.5   & 65.8 /   48.3    &  72.7 / 54.5    &    63.0 / 48.0   &    72.6 / 52.1  &  70.4 / 53.9  \\
\bottomrule
\end{tabular}
}
\caption{Average F$_1$ and Exact Match results for \textbf{pre-trained languages}, on the test set of \textbf{XQuAD} for the \textsc{X-Mod} and \textsc{shared} model variants, pre-trained on the set of 60 languages for 265k update steps. \textbf{Bold} numbers indicate better performance for the respective language.}
\label{tab:xquad_pre-trained}
\end{table}

\begin{table}[t!]
    \centering
        \def\arraystretch{0.87}
\resizebox{0.99\columnwidth }{!}{
\begin{tabular}{lrrr|r}
\toprule
 &  \multicolumn{1}{c}{de}    &     \multicolumn{1}{c}{es} &     \multicolumn{1}{c}{zh}  &   \multicolumn{1}{|c}{\textit{avg}}\\
    & F$_1$ / EM & F$_1$ / EM & F$_1$ / EM & F$_1$ / EM   \\
\midrule
\textsc{X-Mod} & \bf  63.8  / 48.9 & \bf   68.8  / 50.3 &      \bf  61.7     / 36.4  &  \bf 64.8 / 45.2      \\
\textsc{shared}  &  58.9 / 44.1     & 66.7   / 48.3     &  56.5  /    32.2  &   60.7 /   41.5         \\
\bottomrule
\end{tabular}
}

\caption{Average F$_1$ and Exact Match results for \textbf{added languages}, on the test set of \textbf{MLQA} for the \textsc{X-Mod} and \textsc{shared} model variants, pre-trained on the set of 60 languages for 265k update steps. \textbf{Bold} numbers indicate better performance for the respective language.}
\label{tab:mlqa_added}
\end{table}

\begin{table*}[t]
    \centering
        \def\arraystretch{0.87}
\resizebox{0.99\textwidth }{!}{
\begin{tabular}{lr|rrrrrrrrrrrrrrrrrrr|r}
\toprule
 &  en&  af &    ar &    bn &       et &    eu &    fa &    fi &    fr &    hi &    hu &    id &    it &    ka &    ko &    ru &    sw &    ta &   th &    vi & avg \\
\midrule
 \textsc{X-Mod} &  81.4 &  \bf 78.9 &  43.5 & \bf 63.2  & \bf  76.2 & \bf 62.2 & \bf 44.3 & \bf 78.6 & \bf 77.2 & \bf 70.1 & \bf 78.3 &  50.5 & \bf 78.7 & \bf 67.3 & \bf 53.0 & \bf 59.1 & \bf 73.4 &  51.1 &  2.8 & \bf 66.2 & \bf 62.8\\
\textsc{shared}  &  \bf 81.5 &  74.1 & \bf 44.2 &  62.4 &  70.7 &  58.1 &  40.3 &  74.4 &  74.7 &  64.4 &  74.2 & \bf 51.5 &  75.5 &  61.5 &  46.0 &  58.3 &  57.2 & \bf 52.5 & \bf 4.0 &  63.7 & 59.5\\
\bottomrule
\end{tabular}
}
\caption{Average F$_1$ results for \textbf{pre-trained languages}, on the test set of \textbf{NER} for the \textsc{X-Mod} and \textsc{shared} model variants, pre-trained on the set of 60 languages. \textbf{Bold} numbers indicate better performance for the respective language.}
\label{tab:ner_pre-trained}
\end{table*}

\begin{table*}[t]
\centering
\resizebox{\textwidth}{!}{%
\begin{tabular}{lllrcccc}
\toprule
Language & iso & Family &  Script & 13 & 30 & 60 & 75  \\
\midrule
Afrikaans & af & IE:Germanic & Latin   &   &  & \checkmark & \checkmark \\
Albanian & sq & IE:Albanian & Latin  &   & \checkmark & \checkmark & \checkmark\\
Amharic & am & Afro-Asiatic & Amharic  &  & \checkmark & \checkmark \\
Arabic & ar & Afro-Asiatic & Arabic & \checkmark,(+) & \checkmark,(+) & \checkmark,(+) & \checkmark,(+)  \\
Armenian & hy & IE:Armenian & Armenian &   & \checkmark & \checkmark & \checkmark\\
Assamese & as &IE:Iranian & Assamese &  & & & \checkmark \\
Basque &  eu & Isolate & Latin &   & \checkmark & \checkmark & \checkmark\\
Belarusian & be &IE:Slavic & Cyrillic &   &  & \checkmark & \checkmark \\
Bengali & bn &IE:Iranian & Bengali &   &  & \checkmark & \checkmark \\
Bosnian & bs & IE:Slavic & Latin &  & & & \checkmark \\
Breton & br & IE:Celtic & Latin &  & & & \checkmark \\
Bulgarian & bg & IE:Slavic & Cyrillic & + & + & + & + \\
Catalan & ca & IE:Romance & Latin &   &  & \checkmark & \checkmark \\
Chinese & zh & Sino-Tibetan & Chinese & + & + & + & +\\ 
Croatian & hr & IE:Slavic & Latin &   & \checkmark & \checkmark & \checkmark\\
Czech & cs & IE:Slavic & Latin &   & \checkmark & \checkmark & \checkmark \\
Danish & da & IE:Germanic & Latin &   &  & \checkmark & \checkmark \\
Dutch & nl  & IE:Germanic & Latin &   &  & \checkmark & \checkmark \\
English & en & IE:Germanic & Latin  & \checkmark,(+) & \checkmark,(+) & \checkmark,(+) & \checkmark,(+) \\
Estonian & et & Uralic & Latin &   &  & \checkmark & \checkmark \\
Esperanto & eo & Constructed & Latin &   &  & \checkmark & \checkmark \\
Finnish & fi & Uralic  & Latin & \checkmark & \checkmark & \checkmark & \checkmark \\
French & fr  & IE:Romance & Latin & \checkmark,(+) & \checkmark,(+) & \checkmark,(+) & \checkmark,(+) \\
Frisian & fy & IE:Germanic & Latin &  & & & \checkmark \\
Galician & gl & IE:Romance & Latin &   &  & \checkmark & \checkmark \\
Georgian & ka & Kartvelian & Georgian & \checkmark & \checkmark & \checkmark & \checkmark \\
German & de & IE:Germanic & Latin & +,(\checkmark) & +,(\checkmark) & +,(\checkmark) & +,(\checkmark)\\
Greek & el & IE:Hellenic & Greek & +,(\checkmark) & +,(\checkmark) & +,(\checkmark) & +,(\checkmark)\\
Gujarati & gu & IE:Iranian & Gujarati &   &  & \checkmark & \checkmark \\
Hausa & ha & Afro-Asiatic & Latin &   &  & \checkmark & \checkmark \\
Hebrew & he & Afro-Asiatic & Hebrew & +,(\checkmark) & +,(\checkmark) & +,(\checkmark) & +,(\checkmark)\\
Hindi & hi & IE:Iranian& Devanagari & \checkmark,(+) & \checkmark,(+) & \checkmark,(+) & \checkmark,(+)  \\
Hungarian & hu & Uralic & Latin &   & \checkmark & \checkmark & \checkmark\\
Icelandic & is & IE:Germanic & Latin &   &  & \checkmark & \checkmark \\
Indonesian & id & Austronesian & Latin & \checkmark & \checkmark & \checkmark & \checkmark \\
Irish & ga & IE:Celtic & Latin &   &  & \checkmark & \checkmark \\
Italian & it & IE:Romance & Latin &   & \checkmark & \checkmark & \checkmark\\
Japanese & ja & Japonic & Japanese & +,(\checkmark) & +,(\checkmark) & +,(\checkmark) & +,(\checkmark)\\
Javanese & jv & Austronesian & Latin &  & & & \checkmark \\
Kannada & kn & Dravidian & Kannada &  & & & \checkmark \\
Korean & ko & Koreanic & Korean & \checkmark,(+) & \checkmark,(+) & \checkmark,(+) & \checkmark,(+) \\
Kurdish & ku & IE:Iranian & Latin &   &  & \checkmark & \checkmark \\
Latin & la & IE:Romance & Latin &   &  & \checkmark & \checkmark \\
\bottomrule
\end{tabular}

\quad

\begin{tabular}{lllrcccc}
\toprule
Language & iso & Family &Script & 13 & 30 & 60 & 75 \\
\midrule
Latvian & lv &IE:Slavic & Latin &   &  & \checkmark & \checkmark \\
Lithuanian & lt &IE:Slavic & Latin &   & \checkmark & \checkmark & \checkmark\\
Macedonian & mk & IE:Slavic & Cyrillic &   &  & \checkmark & \checkmark \\
Malagasy & mg & Austronesian & Latin &  & & & \checkmark \\
Malay & ms & Austronesian & Latin &   & \checkmark & \checkmark & \checkmark\\
Malayalam & ml & Dravidian & Malayalam &   & \checkmark & \checkmark & \checkmark\\
Marathi & mr & IE:Iranian  & Devanagari &  & & & \checkmark \\
Mongolian & mn & Mongolian & Cyrillic &   & \checkmark & \checkmark & \checkmark\\
Nepali & ne & IE:Iranian & Devanagari &   &  & \checkmark & \checkmark \\
Norwegian  & no & IE:Germanic & Latin &   &  & \checkmark & \checkmark \\
Oriya & or & IE:Iranian  & Odia &  & & & \checkmark \\
Oromo & om & Afro-Asiatic & Ge'ez &  & & & \checkmark \\
Pashto & ps & IE:Iranian &  Arabic &   &  & \checkmark & \checkmark  \\
Persian & fa & IE:Iranian & Arabic &   &  & \checkmark & \checkmark \\
Polish & pl &IE:Slavic  & Latin &   & \checkmark & \checkmark & \checkmark\\
Portuguese & pt & IE:Romance & Latin &   &  & \checkmark & \checkmark \\
Punjabi & pa & IE:Iranian & Gurmukhi &  & & & \checkmark \\
Romanian & ro &IE:Romance & Latin &   & \checkmark & \checkmark & \checkmark\\
Russian & ru & IE:Slavic & Cyrillic & \checkmark,(+) & \checkmark,(+) & \checkmark,(+) & \checkmark,(+) \\
Sanskrit & sa & IE:Iranian & Devanagari &   &  & \checkmark & \checkmark  \\
Scottish Gaelic  & gd & IE:Germanic & Latin &  & & & \checkmark \\
Serbian & sr  & IE:Slavic & Cyrillic &   &  & \checkmark & \checkmark \\
Sindhi & sd & IE:Iranian & Arabic &   &  & \checkmark & \checkmark \\
Sinhala & si & IE:Iranian & Sinhala &   & \checkmark & \checkmark & \checkmark \\
Slovak & sk & IE:Slavic & Latin &   & \checkmark & \checkmark & \checkmark\\
Slovenian & sl & IE:Slavic & Latin &   &  & \checkmark & \checkmark \\
Somali & so & Afro-Asiatic & Latin &   &  & \checkmark & \checkmark \\
Spanish & es & IE:Romance  & Latin & +,(\checkmark) & +,(\checkmark) & +,(\checkmark) & +,(\checkmark)\\
Sundanese & su & Austronesian & Latin   &  & & & \checkmark \\
Swahili & sw & Niger-Congo & Latin  & \checkmark & \checkmark & \checkmark & \checkmark \\
Swedish & sv& IE:Germanic & Latin &   & \checkmark & \checkmark & \checkmark\\
Tagalog & tl & Austronesian & Latin &   & \checkmark & \checkmark & \checkmark\\
Tamil & ta & Dravidian & Tamil & \checkmark & \checkmark & \checkmark & \checkmark \\
Telugu & te & Dravidian  & Telugu &   &  & \checkmark & \checkmark \\
Thai & th &Kra-Dai & Thai & \checkmark,(+) & \checkmark,(+) & \checkmark,(+) & \checkmark,(+) \\
Turkish & tr & Turkic & Latin & +,(\checkmark) & +,(\checkmark) & +,(\checkmark) & +,(\checkmark)\\
Ukrainian & uk & IE:Slavic  & Cyrillic & +,(\checkmark) & +,(\checkmark) & +,(\checkmark) & +,(\checkmark)\\
Urdu & ur & IE:Iranian & Arabic & +,(\checkmark) & +,(\checkmark) & +,(\checkmark) & +,(\checkmark)\\
Vietnamese & vi & Austroasiatic & Latin & \checkmark,(+) & \checkmark,(+) & \checkmark,(+) & \checkmark,(+) \\
Welsh & cy & IE:Celtic & Latin &   &  & \checkmark & \checkmark \\
Xhosa & xh & Niger-Congo & Latin &  & & & \checkmark \\
Yiddish & yi & IE:Germanic & Hebrew &  & & & \checkmark \\
\\
\bottomrule
\end{tabular}
}
\caption{List of languages we  pre-train \checkmark on or add + in the different sets (13, 30, 60, 75). ($\cdot$) indicates the respectively different pre-training/added languages of models 1 and 2 as described in \S\ref{sec:added_langs} and Table~\ref{tab:flipped-languagelist}. IE stands for Indo-European.}
\label{table:appendix_list_of_languages}
\end{table*}

\end{document}